\documentclass{llncs}

\usepackage{amsmath}
\usepackage{amssymb}
\usepackage{hyperref}
\usepackage[caption]{subfig} 
\usepackage{xspace}
\usepackage{tikz}

\usepackage{algorithm}
\usepackage{graphicx}
\usepackage{float}
\usepackage[noend]{algpseudocode}
\usetikzlibrary{positioning}
\usetikzlibrary{calc}
\usepackage[draft]{fixme}

\usepackage{chapterbib}
\usepackage[numbers, sectionbib]{natbib}

\begin{document}
\renewcommand{\bibname}{References}



\title{A fair comparison of many max-tree computation algorithms
 \newline \textit{(Extended version of the \newline paper submitted to ISMM 2013)}}

\author{Edwin Carlinet\inst{1} \and Thierry G\'eraud\inst{1} }
\institute{
EPITA Research and Development Laboratory (LRDE)\\
\email{edwin.carlinet@lrde.epita.fr, thierry.geraud@lrde.epita.fr}
}


\maketitle

\begin{abstract}
With the development of connected filters for the last decade, many
algorithms have been proposed to compute the max-tree. Max-tree allows
to compute the most advanced connected operators in a simple
way. However, no fair comparison of algorithms has been proposed yet
and the choice of an algorithm over an other depends on many
parameters. Since the need of fast algorithms is obvious for
production code, we present an in depth comparison of five algorithms
and some variations of them in a unique framework. Finally, a decision
tree will be proposed to help user choose the right algorithm with
respect to their data.
\end{abstract}




\section{Introduction}
In mathematical morphology, connected filters are those that modify an
original signal by only removing connected components, hence those
that preserve image contours. At the early stage, they were mostly
used for image filtering \citep{vincent1993grayscale,
  salembier1995flat}. Breaking came from max and min-tree as
hierarchical representations of connected components and from an
efficient algorithm able to compute them
\citep{salembier1998antiextensive}. Since then, usage of these trees
has soared for more advanced forms of filtering: based on attributes
\citep{jones1999connected}, using new filtering strategies
\citep{salembier1998antiextensive, urbach2002shape}, allowing new
types of connectivity.  They are also a base for other image
representations, in \citep{monasse2000fast} a tree of shapes is
computed from a merge of min and max trees, in
\citep{xu2012morphological} a component tree is computed over the
attributes values of the max-tree. Max-trees have been used in many
applications: computer vision through motion extraction
\citep{salembier1998antiextensive}, features extraction with MSER
\citep{matas2004robust}, segmentation, 3D visualization
\citep{meijster2002interactive}. With the increase of applications
comes an increase of data type to process: 12-bit image in medical
imagery \citep{meijster2002interactive}, 16-bit or float image in
astronomical imagery \citep{berger2007effective}, and even
multivariate data with special ordering relation
\citep{perret2010connected}. With the improvement of optical sensors,
images are getting bigger (so do image data sets) which urge the need
of fast algorithms. Many algorithms have been proposed to compute the
max-tree efficiently but only partial comparisons have been
proposed. Moreover, some of them are dedicated to particular task
(e.g., filtering) and are unusable for other purpose. We provide in
this paper a full and fair comparison of state-of-the-art max-tree
algorithms in a unique framework i.e. same architecture, same language
(C++) and same outputs.

\section{A tour of max-tree: definition, representation and algorithms}
\subsection{Basic notions for max-tree}
Let $ima: \Omega \rightarrow V$ an image on regular domain $\Omega$,
having values on totally preordered set $(V, <)$ and let
$\mathcal{N}$ a neighborhood on $\Omega$. Let $\lambda \in V$, we note
$[ima \le \lambda]$ the set $\{p \in \Omega, ima(p) \le
\lambda\}$. Let $X \subset \Omega$, we note $CC(X) \subset
\mathcal{P}(\Omega)$ the set of connected components of $X$ w.r.t the
neighborhood $\mathcal{N}$. $CC([ima = \lambda]), \lambda \in V \}$
are \textit{level components} and $\{ CC([ima \ge \lambda]), \lambda
\in V \}$ (resp. $\le$) is the set of upper components (resp. lower
components). The latter endowed with the inclusion relation form a
tree called the max-tree (resp. min-tree). The peak component of $p$
at level $\lambda$ noted $P_p^\lambda$ is the upper component $X \in
CC([ima \ge \lambda])$ such that $p \in X$.

\subsection{Max-tree representation}

\citet{berger2007effective} rely on a simple and effective encoding of
component-trees using an image that stores the \textit{parent}
relationship that exists between components. A connected component is
represented by a single point called the \textit{canonical element}
\citep{berger2007effective,najman2006building} or \textit{level
  root}. Let two points $p,q \in \Omega$, and $p_r$ the root of the
tree, we say that $p$ is canonical if $p = p_r$ or $ima(parent(p))
< ima(p)$. A $parent$ image shall verify the following three properties:
1) $parent(p) = p \Rightarrow p = p_r$ - the root points to itself and
it is the only point verifying this property - 2) $ima(parent(p)) \le
ima(p)$ and 3) $parent(p)$ is canonical.

\newsavebox{\tempbox}
\sbox{\tempbox}{\begin{tikzpicture}[yscale=-1, scale=0.6]
\tikzset{lbl/.style={shift={(-6pt,6pt)}, execute at begin node={\it \scriptsize} }};
\draw[shift={(-0.5,-0.5)}] (0,0) grid (3,3);
\begin{scope}[every node/.style={inner sep=2pt}]
\node[lbl] at (0,0) {A};
\node[lbl] at (0,1) {D};
\node[lbl] at (0,2) {G};
\node[lbl] at (1,0) {B};
\node[lbl] at (1,1) {E};
\node[lbl] at (1,2) {H};
\node[lbl] at (2,0) {C};
\node[lbl] at (2,1) {F};
\node[lbl] at (2,2) {I};
\node (A) at (0,0) {B};
\node (D) at (0,1) {E};
\node (G) at (0,2) {E};
\node (B) at (1,0) {E}; 
\node (E) at (1,1) {F}; 
\node (H) at (1,2) {E}; 
\node (C) at (2,0) {B}; 
\node (F) at (2,1) {F};   
\node (I) at (2,2) {E};
\end{scope}
\node at (1,2.8) {$parent$};

\foreach \x/\y in {A/B, B/E, C/B, D/E, E/F, G/E, H/E, I/E}
	\draw[-latex, gray]  (\x) -- (\y);
\draw[-latex, gray]  (F) .. controls ++(0.8,0.3)  and ++(0.8,-0.3) ..  (F);

\begin{scope}[xshift=-0.5cm]
\node at (-0.9,3.75) {$S$};
\draw[shift={(-0.5,3.5)}] (0,0) grid[step=0.5] (4.5,0.5);
\foreach \x/\xtext in {0/F, 0.5/E, 1/B, 1.5/H, 2/I, 2.5/D, 3/C, 3.5/G, 4/A}
	\node[shift={(-4pt, -4pt)}, inner sep=1pt] (\xtext) at (\x,3.5) {\small \xtext};
\foreach \x/\y in {A/B, B/E, C/B}
	\draw[-latex, gray] (\x.south)  -| ++(0,0.3) -| ($ (\y.south east)  - (0.1,0) $); 
\foreach \x/\y in {D/E, E/F, G/E, H/E, I/E}
	\draw[-latex, gray]  (\x.north)  -| ++(0,-0.3) -| ($ (\y.north east)  - (0.1,0) $); 
\end{scope}
\end{tikzpicture}}
\begin{figure}[htpb]
\centering
\subfloat[Original image $ima$]{
  \vbox to \ht\tempbox{\vfill \hbox{\begin{tikzpicture}[yscale=-1, scale=0.6]
\tikzset{lbl/.style={shift={(-6pt, 6pt)}, execute at begin node={\it \scriptsize} }};
\draw[shift={(-0.5,-0.5)}] (0,0) grid (3,3);
\node at (0,0) {15}; \node[lbl] at (0,0) {A};
\node at (0,1) {12}; \node[lbl] at (0,1) {D};
\node at (0,2) {16}; \node[lbl] at (0,2) {G};
\node at (1,0) {13}; \node[lbl] at (1,0) {B};
\node at (1,1) {12}; \node[lbl] at (1,1) {E};
\node at (1,2) {12}; \node[lbl] at (1,2) {H};
\node at (2,0) {16}; \node[lbl] at (2,0) {C};
\node at (2,1) {10};   \node[lbl] at (2,1) {F};
\node at (2,2) {14}; \node[lbl] at (2,2) {I};
\end{tikzpicture}} \vfill}
}
\subfloat[Max-tree of $ima$]{
  \vbox to \ht\tempbox{\vfill \hbox{\begin{tikzpicture}[yscale=-1, scale=0.6]
\foreach \x in {0,1,...,3}{
	\pgfmathtruncatemacro{\v}{10+2*\x}
	\node at (-3,\x) {\v};
	\draw[dashed, gray] (-2.5,\x) -- ++(4.8,0);
}

\tikzset{notcan/.style={fill=white, draw=gray, inner sep=0.2pt, execute at begin node={\it \scriptsize}}}
\path[every node/.style={circle, fill=white, draw, inner sep=0.05cm, inner sep=1pt, text width=7pt, text centered}]
node (F) at (0,0) {F} [grow=up, level distance=1cm]
	child[grow=up]{ node (E) {E}
		child[grow=up, level distance=0] { node[draw=none, opacity=0] {E}
			child[level distance=2cm] { node {G} }
			child[level distance=1cm] { node {I} }
			child[level distance=0.5cm]{ node {B}
				child[level distance=1cm] { node {A} }
				child[level distance=1.5cm] { node {C} }
			}
		}
	}
	node (H) at (1.1,1.2) [notcan] {H}
	node (D) at (2,1.2) [notcan] {D}
;
\draw (H) to[bend left] (E);
\draw (D) to[bend left] (E);

\end{tikzpicture}} \vfill }
}
\subfloat[Max-tree representation]{\usebox{\tempbox}}
\caption{Representation of a max-tree with a parent image and an array.}
\label{fig:maxtree-repr}
\end{figure}
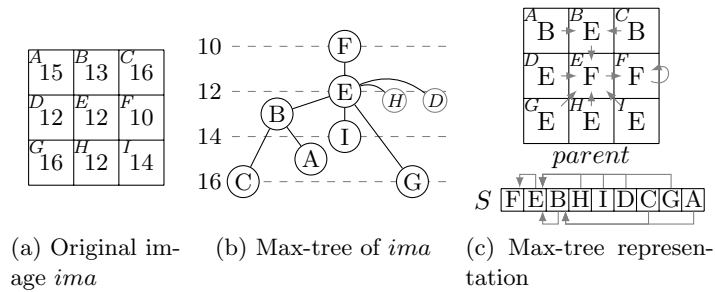

Furthermore, this representation requires an extra vector $S:
\mathbb{N} \rightarrow \Omega$ of points that orders the nodes
downward. Thus, $S$ must verifies $\forall i,j \in \mathbb{N} \; i < j
\Rightarrow S[j] \ne \mathit{parent}(S[i])$. Ordering $S$ vector
allows to traverse the tree upward or downward without storing
children of each node. \autoref{fig:maxtree-repr} shows an example of
such a representation of a max-tree.  This representation only
requires $2.n.I$ bytes memory space where $n$ is the number of pixels
and $I$ the size in bytes of an integer (points are positive
offsets in a pixel buffer). The algorithms we compare have all been
modified to output such a tree encoding.



\subsection{Attribute filtering and reconstruction}
A classical approach for object detection and filtering is to compute
some features called attributes on max-tree nodes. An usual attribute
is the number of pixels in components. Followed by a filtering, it
leads to the well-known area opening. More advanced attributes have
been used like elongation, moment of inertia
\citep{wilkinson2001shape} or even mumford-shah like energy
\citep{xu2012morphological}.

Many max-tree algorithms only construct the $parent$ image but do not
care about $S$ construction \citep{wilkinson2008concurrent,
  matas2004robust}, they output incomplete information. We require
that max-tree algorithms give a ``usable'' tree, i.e., a tree that can
be traversed upwards and downwards, that allows attribute
computation and non-trivial filtering.  The rational behind this
requirement is that, for some applications, filtering parameters are
not known yet when building the tree (e.g., for interactive
visualization \citep{meijster2002interactive}).  In the algorithms we
compare in this paper, no attribute computation nor filtering are
performed during tree construction for clarity reasons; yet they can
be augmented to compute attribute and filtering at the same time.
Algorithm \autoref{alg:attribute} provides an implementation of
attribute computation and direct-filtering with the representation.
$\hat{f}: \Omega \times V \rightarrow \mathcal{A}$ is an application
that projects a pixel $p$ and its value $ima(p)$ in the attribute space
$\mathcal{A}$. $\hat{+}: \mathcal{A} \times \mathcal{A} \rightarrow
\mathcal{A}$ is an associative operator used to merge attributes of
different nodes. \Call{compute-attribute}{} starts with computing
attribute of each singleton node and merge them from leaves toward
root. Note that this simple code relies on the fact that a node
receives all information from its children before passing its
attribute to the parent. Without any ordering on $S$, it would not have
been possible.  \Call{direct-filter}{} is an implementation of direct
filtering as explained in \citep{salembier1998antiextensive} that
keeps all nodes passing a criterion $\lambda$ and lowers nodes that
fails to the last ancestor ``alive''. This implementation has to be
compared with the one in \citep{wilkinson2008concurrent} that only
uses $parent$. This one is shorter, faster and clearer above all.

\begin{algorithm}
\caption{Computation of attributes and filtering.}
\label{alg:attribute}
\hspace{-10pt}
\begin{minipage}[t]{0.5\linewidth}
\begin{algorithmic}
  \Function{\\compute-attribute}
           {$S$, $parent$, $ima$}
  \State $p_{root} \gets S[0]$
  \ForAll {$p \in S$}
  \State $attr(p) \gets \hat{f}(p, ima(p))$
  \EndFor \vspace{-3pt}
  \ForAll {$p \in S$ backward, $p \ne p_{root}$}
  \State $q \gets parent(p)$
  \State $attr(q) \gets attr(q) \hat{+} attr(p)$
  \EndFor \vspace{-3pt}
  \State \Return $attr$
  \EndFunction
\end{algorithmic}
\end{minipage}
\hspace{-15pt}
\begin{minipage}[t]{0.55\linewidth}
\begin{algorithmic}
  \Function{\\direct-filter}{$S$, $parent$, $ima$, $attr$}
  \State $p_{root} \gets S[0]$
  \If { $attr(p_{root}) < \lambda$ } $out(p_{root}) = 0$
  \Else \, $ out(p_{root}) = ima(p_{root})$
  \EndIf
  \ForAll {$p \in S$ forward}
  \State $q \gets parent(p)$
  \If {$ima(q) = ima(p)$}
  \State $out(p) = out(q)$ \Comment {$p$ not canonical}
  \ElsIf {$attr(p) < \lambda$}
  \State $out(p) \gets out(q)$ \Comment{Criterion failed}
  \Else
  \State $out(p) \gets ima(p)$ \Comment {Criterion pass}
  \EndIf
  \EndFor \vspace{-3pt}
  \State \Return $out$
  \EndFunction
\end{algorithmic}
\end{minipage}
\end{algorithm}

\section{Max-tree algorithms}
Max-tree algorithms can be classified in three classes:
\begin{description}
  \item[Immersion algorithms.] It starts with building $N$ disjoints
    singleton for each pixel and sort them according to their gray
    value. Then, disjoint sets merge to form a tree using Tarjan's
    union-find algorithm \citep{tarjan1975efficiency}.
  \item[Flooding algorithms.] A first scan allows to retrieve the root
    which is a pixel at lowest level in the image. Then, it performs a
    propagation by flooding firsts the neighbor at highest
    level i.e. a depth first propagation.
    \citep{salembier1998antiextensive,wilkinson2011fast}.
  \item[Merge-based algorithms.] They divide an image in blocks and
    compute the max-tree on each sub-image using another max-tree
    algorithm. Sub max-trees are then merged to form the tree of the
    whole image. Those algorithms are well-suited for parallelism
    using a \textit{map-reduce} approach
    \citep{wilkinson2008concurrent}. When blocks are image lines, a
    dedicated 1D max-tree algorithm can be used
    \citep{matas2008parallel,menotti20071d}.
\end{description}

\subsection{Immersion algorithms}
\citet{berger2007effective} and \citet{najman2006building} proposed
two algorithms based on Tarjan's union-find. They consist in tracking
disjoints connected components and merge them in bottom up
fashion.  First, pixels are sorted in an array $S$ where each pixel $p$
represent the singleton set $\{p\}$.  Then, we process pixels of $S$
in backward order. When a pixel $p$ is processed, it looks for already
processed neighbor ($\mathcal{N}(p)$) and merges with neighboring
connected components to form a new connected set rooted in $p$. The
merging process consists in updating the \textit{parent} pointer of
neighboring component roots toward $p$. Thus, the union-find relies on
three processes:
\begin{itemize}
\item \verb|make-set(parent, x)| that builds the singleton set $\{x\}$,
\item \verb|find-root(parent, x)| that finds the root of the component that
  contains $x$,
\item \verb|merge-set(parent, x, y)| that merges components rooted in $x$
  and $y$ and set $x$ as the new root.
\end{itemize}
Based on the above functions, a simple max-tree algorithm is given below:

\begin{algorithmic}
\Procedure{Maxtree}{$ima$}
\State $S \gets $ sorts pixels increasing
\ForAll{$p \in S$ backward}
\State $\textrm{make-set}(parent, p)$
\ForAll{$n \in \mathcal{N}_p$ processed}
\State $r \gets \textrm{find-root}(parent, n)$
\If {$r \ne p$}
\State $\textrm{merge-set}(parent, p, r)$
\EndIf
\EndFor
\EndFor
\EndProcedure
\end{algorithmic}

\verb|find-root| is a $O(n)$ function that makes the above procedure a
$O(n^2)$ algorithm. \citet{tarjan1975efficiency} discussed two
important optimizations for his algorithm to avoid a quadratic
complexity:
\begin{description}
\item Root path compression. When \textit{parent} is traversed to get the root
  of the component, points of the path used to find the root
  collapse to the actual root the component. However, path compression
  should not be applied on $parent$ image because it removes the
  hierarchical structure of the tree. As consequence, we apply path
  compression on an intermediate image $zpar$ that stores the root of
  disjoints components. Path compression bounds union-find complexity
  to $O(n \log n)$ and has been applied in \citep{berger2007effective}
  and \citep{najman2006building}.
\item Union-by-rank. When merging two components $A$ and $B$, we have to select
  one the roots to represent the newly created component. If $A$
  has a \textit{rank} greater than $B$ then $root_A$ is selected as the new
  root, $root_B$ otherwise. When rank matches the depth of trees, it
  enables tree balancing and guaranties a $O(n \log n)$ complexity for
  union-find. When used with path compression, it allows to compute
  the max-tree in quasi-linear time ($O(n.\alpha(n))$ where
  $\alpha(n)$ is the inverse of Ackermann function which is very
  low-growing). Union-by-rank has been applied in
  \citep{najman2006building}.
\end{description}

Note that \textit{parent} and \textit{zpar} encode two different
things, \textit{parent} encodes the max tree while \textit{zpar}
tracks disjoints set of points and also use a tree. Thus,
union-by-rank and root path compression shall always be applied on
\textit{zpar} but never on \textit{parent}.

Algorithm \ref{alg:berger} is the union-find based max-tree algorithm as
proposed by \citet{berger2007effective}. It starts with sorting pixels
that can be done with a counting sort algorithm for low-quantized data or
with a radix sort-based algorithm for high quantized
data\citep{andersson1995sorting}.Then it annotates all pixels as
\textit{unprocessed} with $-1$ (in standard implementations pixel are
positive offsets in a pixel buffer). Later in the algorithm, when a
pixel $p$ is processed it becomes the root of the component i.e $parent(p) =
p$ with $p \ne -1$, thus testing $parent(p) \ne -1$ stands for
\textit{is $p$ already processed}. Since $S$ is processed in reverse
order and \texttt{merge-set} sets the root of the tree to the current
pixel $p$ ($parent(r) \leftarrow p$), it ensures that the parent $p$ will be
seen before its child $r$ when traversing $S$ in the direct order.

\begin{algorithm}[htb]
\caption{Union find without union-by-rank}
\label{alg:berger}
\begin{algorithmic}
\Function{Find-root}{$par$, $p$}
\State \textbf{if} $par(p) \ne p$ \textbf{ then } $par(p) \gets \Call{find-root}{par, par(p)}$
\State \Return $par(p)$
\EndFunction
\Function{Maxtree}{$ima$}
\ForAll{$p$} $parent(p) \gets -1$ \EndFor
\State $S \gets $ sorts pixels increasing
\ForAll{$p \in S$ backward}
\State $parent(p) \gets p$; $zpar(p) \gets p$  \Comment{make-set}
\ForAll{$n \in \mathcal{N}_p$ such that $parent(n) \ne -1$}
\State $r \gets \Call{find-root}{zpar, n}$
\If {$r \ne p$}
\State $zpar(r) \gets p$; $parent(r) \gets p$ \Comment{merge-set}
\EndIf
\EndFor
\EndFor
\State \Call{Canonize}{$parent$, $S$}
\State \Return $(parent, S)$
\EndFunction
\end{algorithmic}
\end{algorithm}

\subsubsection{Union-by-rank}

Algorithm \ref{alg:bergerrank} is similar to algorithm \ref{alg:berger} but
augmented with union-by-rank. It first introduces a new image
$rank$. The \texttt{make-set} step creates a tree with a single node,
thus with a rank set to 0. The $rank$ image is then used when merging
two connected set in $zpar$. Let $z_p$ the root of the connected
component of $p$, and $z_n$ the root of connected component of $n \in
\mathcal{N}(p)$. When merging two components, we have to decide
which of $z_p$ or $z_n$ becomes the new root w.r.t their
rank. If $rank(z_p) < rank(z_n)$, $z_p$ becomes the root, $z_n$
otherwise. If both $z_p$ and $z_n$ have the same rank then we can
choose either $z_p$ or $z_n$ as the new root, but the rank should be
incremented by one. On the other hand, the relation $parent$ is
unaffected by the union-by-rank, $p$ becomes the new root whatever the
rank of $z_p$ and $z_n$. Whereas without balancing the root of any
point $p$ in $zpar$ matches the root of $p$ in parent, this is not the
case anymore. For every connected components we have to keep a
connection between the root of the component in $zpar$ and the root of
max-tree in $parent$. Thus, we introduce an new image $repr$ that
keeps this connection updated.
\begin{algorithm}[htb]
\caption{Union find with union-by-rank}
\label{alg:bergerrank}
\begin{algorithmic}
\Procedure{Maxtree}{$ima$}
\ForAll{$p$} $parent(p) \gets -1$ \EndFor
\State $S \gets $ sorts pixels increasing
\ForAll{$p \in S$ backward}
\State $parent(p) \gets p$; $zpar(p) \gets p$ \Comment{make-set}
\State $rank(p) \gets 0$; $repr(p) \gets p$
\State $z_p \gets p$
\ForAll{$n \in \mathcal{N}_p$ such that $parent(n) \ne -1$}
\State $z_n \gets \Call{find-root}{zpar, n}$
\If {$z_n \ne z_p$}
\State $parent(repr(z_n)) \gets p$
\If{ $rank(z_p) < rank(z_n)$ } $swap(z_p, z_n)$ \EndIf
\State $zpar(z_n) \gets z_p$  \Comment{merge-set}
\State $repr(z_p) \gets  p$
\If {$rank(z_p) = rank(z_n)$}
\State $rank(z_p) \gets rank(z_p)+1$
\EndIf
\EndIf
\EndFor
\EndFor
\State \Call{Canonize}{$parent$, $S$}
\State \Return $(parent, S)$
\EndProcedure
\end{algorithmic}
\end{algorithm}

\begin{figure}[htb]
\centering
\subfloat[]{\usetikzlibrary{positioning}
\begin{tikzpicture}[baseline, scale=0.6]
\tikzset{lbl/.style={shift={(-4pt,4pt)}, gray, execute at begin node={\it \scriptsize} }};
\tikzset{every node/.style={circle, draw}};
\tikzset{level distance=1.1cm}
\tikzset{sibling distance=1cm}

\draw (-1.5,-2.7)  node[draw=none, above right=50pt and 0.1] {$zpar$}  rectangle ++(4.9,4);
\draw (3.6,-2.7)  rectangle ++(4.9,4) node[draw=none, below left=-5pt and 0.1] {$parent$};
\begin{scope}[every node/.style={draw=none}, yshift=100pt, scale=0.8, yscale=-1]
\small
\draw[shift={(-0.5,-0.5)}] (0,0) grid (3,3);
\node[above] at (1,-0.5) {$zpar$};
\node at (0,0) {B}; \node[lbl] at (0,0) {A};
\node at (0,1) {}; \node[lbl] at (0,1) {D};
\node at (0,2) {H}; \node[lbl] at (0,2) {G};
\node at (1,0) {B}; \node[lbl] at (1,0) {B};
\node[fill=gray!20] at (1,1) {B}; \node[lbl] at (1,1) {E};
\node at (1,2) {H}; \node[lbl] at (1,2) {H};
\node at (2,0) {B}; \node[lbl] at (2,0) {C};
\node at (2,1) {};   \node[lbl] at (2,1) {F};
\node at (2,2) {H}; \node[lbl] at (2,2) {I};
\end{scope}

\begin{scope}[every node/.style={draw=none}, xshift=77pt, yshift=100pt, scale=0.8, yscale=-1]
\small
\draw[shift={(-0.5,-0.5)}] (0,0) grid (3,3);
\node[above] at (1,-0.5) {$repr$};
\node at (0,0) {A}; \node[lbl] at (0,0) {A};
\node at (0,1) {}; \node[lbl] at (0,1) {D};
\node at (0,2) {G}; \node[lbl] at (0,2) {G};
\node at (1,0) {E}; \node[lbl] at (1,0) {B};
\node[fill=gray!20] at (1,1) {E}; \node[lbl] at (1,1) {E};
\node at (1,2) {H}; \node[lbl] at (1,2) {H};
\node at (2,0) {C}; \node[lbl] at (2,0) {C};
\node at (2,1) {};   \node[lbl] at (2,1) {F};
\node at (2,2) {I}; \node[lbl] at (2,2) {I};
\end{scope}

\begin{scope}[every node/.style={draw=none}, xshift=155pt, yshift=100pt, scale=0.8, yscale=-1]
\small
\draw[shift={(-0.5,-0.5)}] (0,0) grid (3,3);
\node[above] at (1,-0.5) {$parent$};
\node at (0,0) {B}; \node[lbl] at (0,0) {A};
\node at (0,1) {}; \node[lbl] at (0,1) {D};
\node at (0,2) {H}; \node[lbl] at (0,2) {G};
\node at (1,0) {E}; \node[lbl] at (1,0) {B};
\node[fill=gray!20] at (1,1) {E}; \node[lbl] at (1,1) {E};
\node at (1,2) {H}; \node[lbl] at (1,2) {H};
\node at (2,0) {B}; \node[lbl] at (2,0) {C};
\node at (2,1) {};   \node[lbl] at (2,1) {F};
\node at (2,2) {H}; \node[lbl] at (2,2) {I};
\end{scope}

\begin{scope}[inner sep=1pt]
\small
\node (z1) at (0,0) {B} 
	child { node {C} }
	child { node {A} }
	child { node (e) { E} }
;

\node (z2) at (2.5,0) {H}
	child { node {G} }
	child { node[text width=0.2cm, text centered] {I} }
;
\node[draw=none, blue, inner sep=0] (p) [above right=-2pt of e] {$p$};
\node[draw=none, blue, inner sep=0] (zp) [above right=-2pt of z1] {$z_p$}; 
\node[draw=none, blue, inner sep=0] (r) [above right=-2pt of z2] {$z_n$};
\node[draw=none, blue, inner sep=0] (n) [left=0pt of z2] {$n$};
\draw[dashed, -latex, blue] (p) -- (zp); 
\draw[dashed, -latex, blue] (n) .. controls ++(0,0.5) .. (r) ;

\draw[blue,-latex, dashed] (-1.3,-2.5) -- ++(0.5,0) node[draw=none, blue, right]  {\small \it find-root}; 

\node (r1) at (4.7,0) {E}
	child { node {B}
		child { node {C} }
		child { node {A} }
	}
;
\node (r2) at (6.7,0) {H}
	child { node {G} }
	child { node[text width=0.2cm, text centered] {I} }
;
\node[draw=none, blue, inner sep=0] (zp) [above right=0pt of r1] {$p$};
\node[draw=none, blue, inner sep=0] (zn) [above right=0pt of r2] {$r$};

\draw[->, blue] (z1) to[bend left] node[draw=none, above=-10pt] {\scriptsize $repr[z_p]$}  (r1);
\draw[->, blue] (z2) to[bend left] node[draw=none, above=-10pt] {\scriptsize $repr[z_n]$}  (r2);
\end{scope}
\end{tikzpicture}}
\subfloat[]{\usetikzlibrary{positioning}
\begin{tikzpicture}[baseline, scale=0.6]
\tikzset{lbl/.style={shift={(-4pt,4pt)}, gray, execute at begin node={\it \scriptsize} }};
\tikzset{every node/.style={circle, draw}};
\tikzset{level distance=1.1cm}
\tikzset{sibling distance=1cm}

\draw (-1,-2.7)  node[draw=none, above right=-6pt and 45pt] {$zpar$}  rectangle ++(4.4,4);
\draw (3.6,-2.7)  node[draw=none, above right=-8pt and 45pt] {$parent$}  rectangle ++(4.4,4);
\begin{scope}[every node/.style={draw=none}, yshift=100pt, scale=0.8, yscale=-1]
\small
\draw[shift={(-0.5,-0.5)}] (0,0) grid (3,3);
\node[above] at (1,-0.5) {$zpar$};
\node at (0,0) {B}; \node[lbl] at (0,0) {A};
\node at (0,1) {}; \node[lbl] at (0,1) {D};
\node at (0,2) {H}; \node[lbl] at (0,2) {G};
\node at (1,0) {B}; \node[lbl] at (1,0) {B};
\node[fill=gray!20] at (1,1) {B}; \node[lbl] at (1,1) {E};
\node at (1,2) {B}; \node[lbl] at (1,2) {H};
\node at (2,0) {B}; \node[lbl] at (2,0) {C};
\node at (2,1) {};   \node[lbl] at (2,1) {F};
\node at (2,2) {H}; \node[lbl] at (2,2) {I};
\end{scope} 

\begin{scope}[every node/.style={draw=none}, xshift=77pt, yshift=100pt, scale=0.8, yscale=-1]
\small
\draw[shift={(-0.5,-0.5)}] (0,0) grid (3,3);
\node[above] at (1,-0.5) {$repr$};
\node at (0,0) {A}; \node[lbl] at (0,0) {A};
\node at (0,1) {}; \node[lbl] at (0,1) {D};
\node at (0,2) {G}; \node[lbl] at (0,2) {G};
\node at (1,0) {E}; \node[lbl] at (1,0) {B};
\node[fill=gray!20] at (1,1) {E}; \node[lbl] at (1,1) {E};
\node at (1,2) {H}; \node[lbl] at (1,2) {H};
\node at (2,0) {C}; \node[lbl] at (2,0) {C};
\node at (2,1) {};   \node[lbl] at (2,1) {F};
\node at (2,2) {I}; \node[lbl] at (2,2) {I};
\end{scope}

\begin{scope}[every node/.style={draw=none}, xshift=155pt, yshift=100pt, scale=0.8, yscale=-1]
\small
\draw[shift={(-0.5,-0.5)}] (0,0) grid (3,3);
\node[above] at (0.5,-0.5) {$parent$};
\node at (0,0) {B}; \node[lbl] at (0,0) {A};
\node at (0,1) {}; \node[lbl] at (0,1) {D};
\node at (0,2) {H}; \node[lbl] at (0,2) {G};
\node at (1,0) {E}; \node[lbl] at (1,0) {B};
\node[fill=gray!20] at (1,1) {E}; \node[lbl] at (1,1) {E};
\node at (1,2) {E}; \node[lbl] at (1,2) {H};
\node at (2,0) {B}; \node[lbl] at (2,0) {C};
\node at (2,1) {};   \node[lbl] at (2,1) {F};
\node at (2,2) {H}; \node[lbl] at (2,2) {I};
\end{scope}

\begin{scope}[inner sep=1pt]
\small
\node (z1) at (1,0.5) {B} 
	child { node {C} }
	child { node {A} }
	child { node { E}  }
	child[edge from parent/.style={draw, blue}]  { node (z2) {H} 
		child[edge from parent/.style={draw, black}, black] { node {G} }
		child[edge from parent/.style={draw, black}, black] { node[text width=0.2cm, text centered] {I} }
		edge from parent node[draw=none, pos=0.4, right=-12pt] {\small $zpar(z_n) \leftarrow z_p$}
	};

\node[draw=none, blue, inner sep=0] (zp) [above right=0pt of z1] {$z_p$};
\node[draw=none, blue, inner sep=0] (zn) [right=0pt of z2] {$z_n$};

\node (r1) at (5.5,0.5) {E} [sibling distance=1.7cm]
	child { node {B} [sibling distance=0.8cm]
		child { node {C} }
		child { node {A} }
	}
	child[edge from parent/.style={draw, blue}] { node (r2) {H} [sibling distance=0.8cm]
		child[edge from parent/.style={draw, black}, black] { node {G} }
		child[edge from parent/.style={draw, black}, black] { node[text width=0.2cm, text centered] {I} }
		edge from parent node[draw=none, pos=0.4,right=-23pt] {\small $parent(r) \leftarrow p$}
	};
\node[draw=none, blue, inner sep=0] (zp) [above right=0pt of r1] {$p$};
\node[draw=none, blue, inner sep=0] (zn) [right=0pt of r2] {$r$};
\end{scope}
\end{tikzpicture}}
\caption{Union-by-rank. (a) State of the algorithm before processing
  the neighbor $H$ from $E$. (b) State of the algorithm after
  processing.}
\label{fig:urank}
\end{figure}
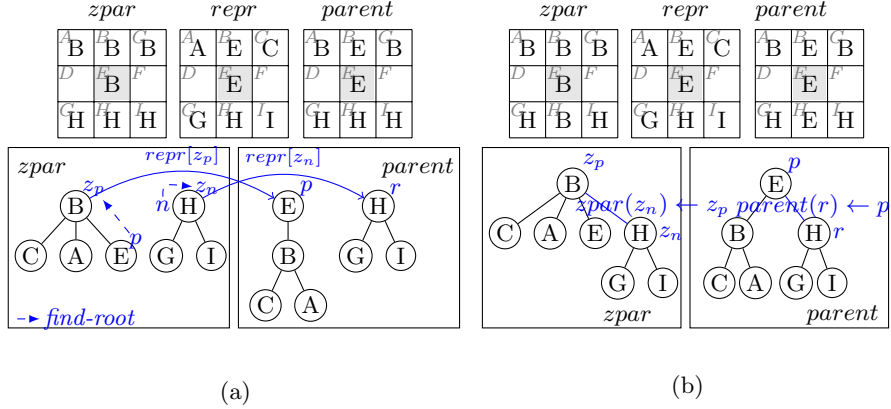

 The union-by-rank technique and structure update are illustrated in
 \autoref{fig:urank}. The algorithm has been running until processing
 $E$ at level $12$, the first neighbor $B$ has already been treated
 and neighbors $D$ and $F$ are skipped because not yet
 processed. Thus, the algorithm is going to process the last neighbor
 $H$. $z_p$ is the root of $p$ in $zpar$ and we retrieve the root
 $z_n$ of $n$ with \texttt{find-root} procedure. Using $repr$ mapping,
 we look up the corresponding point $r$ of $z_n$ in $parent$. The tree rooted
 in $r$ is then merged to the tree rooted in $p$ ($parent$
 merge). Back in $zpar$, components rooted in $z_p$ and $z_n$
 merge. Since they have the same rank, we choose arbitrary $z_p$ to be
 the new root.

Algorithm \ref{alg:bergerrank} is slightly different from the one of
\citet{najman2006building}. They use two union-find structure, one to
build the tree, the other to handle flat zones. In their paper,
\texttt{lowernode[$z_p$]} is an array that maps the root of a
component $z_p$ in \textit{zpar} to a point of current level component
in \textit{parent} (just like \texttt{repr($z_p$)} in our algorithm).
Thus, they apply a second union-find to retrieve the canonical.  This
extra union-find can be avoided because \verb|lowernode[x]| is already
a canonical element, thus $findoot$ on $lowernode(z_p)$ is useless and so
does $parent$ balancing on flat zones.

\subsubsection{Canonization}
Both algorithms call the \Call{Canonize} procedure to ensure
that any node's parent is a canonical node. In algorithm
\ref{alg:canonize}, canonical property is broadcast downward.
$S$ is traversed in direct order such that when processing a pixel $p$,
its parent $q$ has the canonical property that is $parent(q)$ is
a canonical element. Hence, if $q$ and $parent(q)$ belongs to the same
node i.e $ima(q) = ima(parent(q))$, the parent of $p$ is set to the
component's canonical element: $parent(q)$.

\begin{algorithm}[htb]
\caption{Canonization algorithm}
\label{alg:canonize}
\begin{algorithmic}
\Procedure{Canonize}{$ima$, $parent$, $S$}
\ForAll {$p$ in $S$ forward}
\State $q \gets parent(p)$
\If {$ima(q) = ima(parent(q))$}
\State $parent(p) \gets parent(q)$
\EndIf
\EndFor
\EndProcedure
\end{algorithmic}
\end{algorithm}

\subsubsection{Level compression}

Union-by-rank provides time complexity guaranties at the price of
extra memory requirement. When dealing with huge images this results
in a significant drawback (e.g. RAM overflow\ldots). Since the last
point processed always becomes the root, union-find without rank
technique tends to create degenerated tree in flat zones. Level
compression avoids this behavior by a special handling of flat
zones. In algorithm \autoref{alg:lcompress}, $p$ is the point in
process at level $\lambda = ima(p)$, $n$ a neighbor of $p$ already
processed, $z_p$ the root of $P_p^\lambda$ (at first $z_p=p$), $z_n$
the root of $P_n^\lambda$. We suppose $ima(z_p) = ima(z_n)$, thus
$z_p$ and $z_n$ belong to the same node and we can choose any of them
as a canonical element. Normally $p$ should become the root with child
$z_n$ but level compression inverts the relation, $z_n$ is kept as the
root and $z_p$ becomes a child. Since $parent$ may be inverted, $S$
array is not valid anymore. Hence $S$ is reconstructed, as soon as a
point $p$ gets attached to a root node, $p$ will be not be processed
anymore so its inserted in back of $S$.  At the end $S$ only misses
the tree root which is $parent[S[0]]$.

\begin{algorithm}[htb]
\caption{Union find with level compression}
\label{alg:lcompress}
\begin{algorithmic}
\Function{Maxtree}{$ima$}
\ForAll{$p$} $parent(p) \gets -1$ \EndFor
\State $S \gets $ sorts pixels increasing
\State $j = N-1$
\ForAll{$p \in S$ backward}
\State $parent(p) \gets p$; $zpar(p) \gets p$  \Comment{make-set}
\State $z_p = p$
\ForAll{$n \in \mathcal{N}_p$ such that $parent(n) \ne -1$}
\State $z_n \gets \Call{find-root}{zpar, n}$
\If {$z_p \ne z_n$}
\If {$ima(z_p) = ima(z_n)$} $\Call{swap}(z_p, z_n)$ \EndIf
\State $zpar(z_n) \gets z_p$; $parent(z_n) \gets z_p$ \Comment{merge-set}
\State $S[j] \gets z_n$; $j \gets j-1$
\EndIf
\EndFor
\EndFor
\State $S[0] \gets parent[S[0]]$
\State \Call{Canonize}{$parent$, $S$}
\State \Return $(parent, S)$
\EndFunction
\end{algorithmic}
\end{algorithm}

\subsection{Flooding algorithms}

\citet{salembier1998antiextensive} proposed the first efficient
algorithm to compute the max-tree. A propagation starts from the root
that is the pixel at lowest level $l_{min}$. Pixels in the propagation
front are stored in a hierarchical queue that allows a direct access
to pixels at a given level in the queue. The \texttt{flood($\lambda$,
  $r$)} procedure (see algorithm \ref{alg:hqueue}) is in charge of
flooding the peak component $P_r^\lambda$ and building the
corresponding sub max-tree rooted in $r$. It proceeds as follows:
first pixels at level $\lambda$ are retrieved from the queue, their
$parent$ pointer is set to the canonical element $r$ and their
neighbors $n$ are analyzed. If $n$ is not in queue and has not yet
been processed, then $n$ is pushed in the queue for further process
sing and $n$ is marked as processed ($parent(n)$ is set to
\texttt{INQUEUE} which is any value different from -1). If the level
$l$ of $n$ is higher than $\lambda$ then $n$ is in the childhood of
the current node, thus \texttt{flood} is called recursively to flood
the peak component $P_n^l$ rooted in $n$. During the recursive flood,
some points can be pushed in queue between level $\lambda$ and
$l$. Hence, when \texttt{flood} ends, it returns the level $l'$ of
$n$'s parent. If $l' > \lambda$, we need to flood level $l'$ until $l'
\le \lambda$ i.e until there are no more points in the queue above
$\lambda$. Once all pixels at level $\lambda$ have processed, we need
to retrieve the level $lpar$ of parent component and attach $r$ to its
canonical element. A $levroot$ array stores canonical element of each
level component and -1 if the component is empty. Thus we just have to
traverse $levroot$ looking for $lpar = \max \{h < \lambda, levroot[h]
\ne -1 \}$ and set the parent of $r$ to $levroot[lpar]$.  Since the
construction of $parent$ is bottom-up, we can safely insert $p$ in
front of the $S$ array each time $parent(p)$ is set. For a level
component, the canonical element is the last element inserted ensuring
a correct ordering of $S$. Note that the first that gets a the minimum
level of the image is not necessary. Instead, we could have called
\texttt{flood} in \texttt{Max-tree} procedure until the parent level
returned by the function was -1, i.e the last flood call was
processing the root. Anyway, this pass has other advantages for
optimization that will be discussed in the implementation details
section.

\begin{algorithm}[htb]
\caption{\citet{salembier1998antiextensive} max-tree algorithm}
\label{alg:hqueue}
\begin{algorithmic}
\Function{flood}{$\lambda$, $r$}
\While{$hqueue[\lambda]$ not empty}
\State $p \gets \Call{pop}{hqueue[\lambda]}$
\State $parent(p) \gets r$
\If {$p \ne r$}  \Call{insert\_front}{$S$, $p$} \EndIf
\ForAll {$n \in \mathcal{N}(p)$ such that $parent(p) = -1$}
\State $l \gets ima(n)$
\If {$levroot[l] = -1$} $levroot[l] \gets n$ \EndIf
\State $\Call{push}{hqueue[l], n}$
\State $parent(n) \gets $ \texttt{INQUEUE}
\While{$l > \lambda$} \State $l \gets flood(l, levroot[l])$ \EndWhile
\EndFor
\EndWhile
\State \Comment{Attach to parent}
\State $levroot[\lambda] \gets -1$
\State $lpar \gets \lambda - 1$
\While {$lpar \ge 0$ \textbf{and} $levroot[lpar] = -1$} \State $lpar \gets lpar-1$ \EndWhile
\If {$lpar \ne -1$} \State $parent(r) \gets levroot[lpar]$ \EndIf
\State \Call{insert\_front}{$S$, $r$}
\State \Return $lpar$
\EndFunction
  \end{algorithmic}
\end{algorithm}

\begin{algorithm}[htb]
  \ContinuedFloat
  \caption{Salembier maxtree algorithm (continued)}
\begin{algorithmic}
\Function{Max-tree}{ima}
\ForAll {$h$} $levroot[h] \gets -1$ \EndFor
\ForAll {$p$} $parent(p) \gets -1$ \EndFor
\State $l_{min} \gets \min_p ima(p)$
\State $p_{min} \gets \arg \min_p ima(p)$
\State \Call{push}{$hqueue[l_{min}]$, $p_{min}$}
\State $levroot[lmin] \gets p_{min}$
\State \Call{flood}{$l_{min}$, $p_{min}$}
\EndFunction
\end{algorithmic}
\end{algorithm}

\citet{salembier1998antiextensive}'s algorithm was rewritten in a
non-recursive implementation in \citet{hesselink2003salembier} and
later by \citet{nister2008linear} and \citet{wilkinson2011fast}. These
algorithms differ in only two points. First, \citep{wilkinson2011fast}
uses a pass to retrieve the root before flooding to mimics the
original recursive version while \citet{nister2008linear} does
not. Second, priority queues in \citep{nister2008linear} use an
unacknowledged implementation of heap based on hierarchical queues
while in \citep{wilkinson2011fast} they are implemented using a
standard heap (based on comparisons). The algorithm \ref{alg:unrec} is
a code transcription of the method described in
\citet{nister2008linear}. The array $levroot$ in the recursive version
is replaced by a stack with the same purpose: storing the canonical
element of level components. The hierarchical queue $hqueue$ is
replaced by a priority queue $pqueue$ that stores the propagation
front. The algorithm starts with some initialization and choose a
random point $p_{start}$ as the flooding point. $p_{start}$ is
enqueued and pushed on $levroot$ as canonical element. During the
flooding, the algorithm picks the point $p$ at highest level (with the
highest priority) in the queue, and the canonical element $r$ of its
component which is the top of $levroot$ ($p$ is not removed from the
queue). Like in the recursive version, we look for neighbors $n$ of
$p$ and enqueue those that have not yet been seen. If $ima(n) >
ima(p)$, $n$ is pushed on the stack and we immediately flood $n$ (a
\textit{goto} that mimics the recursive call). On the other hand, if
all neighbors are in the queue or already processed then $p$ is
\textit{done}, it is removed from the queue, $parent(p)$ is set its
the canonical element $r$ and if $r \ne p$, $p$ is added to $S$ (we
have to ensure that the canonical element will be inserted last). Once
$p$ removed from the queue, we have to check if the level component
has been fully processed in order to attach the canonical element $r$
to its parent. If the next pixel $q$ has a different level than $p$,
we call the procedure \texttt{ProcessStack} that pops the stack, sets
parent relationship between canonical elements and insert them in $S$
until the top component has a level no greater than $ima(q)$. If the
stack top's level matches $q$'s level, $q$ extends the component so no
more process is needed. On the other hand, if the stack gets empty or
the top level is lesser than $ima(q)$, then $q$ is pushed on the stack
as the canonical element of a new component. The algorithm ends when
all points in queue have been processed, then $S$ only misses the root
of the tree which is the single element that remains on the stack.

\begin{algorithm}[htb]
  \caption{Non-recursive max-tree algorithm \citep{nister2008linear,wilkinson2011fast} }
  \label{alg:unrec}
  \begin{algorithmic}[1]
    \Function{Max-tree}{$ima$}
    \ForAll {$p$} $parent(p) \gets -1$ \EndFor
    \State $p_{start} \gets $ any point in $\Omega$
    \State $\Call{push}{pqueue, p_{start}}$; $\Call{push}{levroot, p_{start}}$
    \State $parent(p_{start}) \gets \texttt{INQUEUE}$
    \Loop
    \State $p \gets \Call{top}{pqueue}$; $r \gets \Call{top}{levroot}$
    \ForAll {$n \in \mathcal{N}(p)$ such that $parent(p) = -1$}
    \State $\Call{push}{pqueue, n}$
    \State $parent(n) \gets \texttt{INQUEUE}$
    \If{$ima(p) < ima(n)$}
    \State $\Call{push}{levroot, n}$
    \State \textbf{goto} 7
    \EndIf
    \EndFor
    \State \texttt{\{ $p$ is done \}}
    \State $\Call{pop}{pqueue}$
    \State $parent(p) \gets r$
    \If{$p \ne r$} \Call{insert\_front}{$S$, $p$} \EndIf
    \EndLoop
    \While {$pqueue$ not empty};
    \State \texttt{\{ all points at current level done ? \}}
    \State $q \gets \Call{top}{pqueue}$
    \If{$ima(q) \ne ima(r)$} \Comment{Attach $r$ to its parent}
    \State \Call{ProcessStack}{$r$, $q$}
    \EndIf
    \EndWhile
    \State \textbf{repeat}
    \State $root \gets \Call{pop}{levroot}$
    \State \Call{insert\_front}{$S$, $root$}
    \EndFunction
  \end{algorithmic}
\end{algorithm}

\begin{algorithm}[H]
  \ContinuedFloat
  \caption{Non-recursive max-tree algorithm (continued)}
  \begin{algorithmic}
    \Procedure{ProcessStack}{$r$, $q$}
    \State $\lambda \gets ima(q)$
    \State $\Call{pop}{levroot}$
    \While{$levroot$ not empty \textbf{and} $\lambda <
      ima(\Call{top}{levroot})$}
    \State \Call{insert\_front}{$S$, $r$}
    \State $r \gets parent(r) \gets \Call{pop}{levroot})$
    \EndWhile
    \If{$levroot$ empty \textbf{or} $ima(\Call{top}{levroot}) \ne \lambda$}
    \State $\Call{push}{levroot, q}$
    \EndIf
    \State $parent(r) \gets \Call{top}{levroot}$
    \State \Call{insert\_front}{$S$, $r$}
    \EndProcedure
  \end{algorithmic}
\end{algorithm}

\subsection{Merge-based algorithms and parallelism}
\begin{figure}[htb]
\centering
\subfloat[]{\begin{tikzpicture}[yscale=0.9]
\draw[thick] (0,0) rectangle (2,2);
\draw[thick, dashed] (0,1) -- ++(2,0);
\draw[densely dashed] (0,1.5) -- ++(2,0);
\draw[densely dashed] (0,0.5) -- ++(2,0);
\draw[dotted] (0,0.25) -- ++(2,0);
\scriptsize
\node at (1,0.12) { $D_5$};
\node at (1,0.35) { $D_4$};
\node at (1,0.7) {\small $D_3$};
\node at (1,1.2) {\small $D_2$};
\node at (1,1.7) {\small $D_1$};

\end{tikzpicture}}
\subfloat[]{\begin{tikzpicture}
\scriptsize
\path [inner sep=1pt, level distance=15pt]
node (root) {} [sibling distance=30pt]
	child { node {} [sibling distance=15pt]
		child {  node[]  (d1) {$D_1$} }
		child {  node[]  (d2) {$D_2$} }
	}
	child { node (r2) {} [sibling distance=15pt]
		child { node[] (d3) {$D_3$} }
		child { node {} [sibling distance=15pt]
			child { node[] (d4) {$D_4$} }
			child { node[]  (d5) {$D_5$} }
		}
	}
;

\draw[red] (d1.south west) |- (root.north east) -- ++(0.5,0) -- ++(0,-0.4)  -| (d2.south east) -- cycle ;
\draw[red] (d3.south west) |- (r2.north east) -- ++(0.5,0) -- ++(0,-0.4)  -| (d4.south east) -| (d3.west) ;
\draw[red] (d5.south west) |- ++(0,0.6) -- ++(0.5,0) -- ++ (0,-0.6)  -- cycle ;

\node[red, below left] at (0,0) {\scriptsize Thread 1};
\node[red, below left] at (1.1,-0.5) {\scriptsize Thread 2};
\node[red, left] at (1.7,-1.3) {\scriptsize Thread 3};
\end{tikzpicture}}
\subfloat[]{
\hspace{-10pt}
\begin{minipage}[b]{6cm}
  \scriptsize
\setlength{\tabcolsep}{3pt}
\begin{tabular}[b]{lll}
Thread 1 & Thread 2 & Thread 3\\
$T_1 \leftarrow f(ima_{|D_1})$ & $T_3 \leftarrow
f(ima_{|D_3})$ & $T_5 \leftarrow f(ima_{|D_5})$ \\
$T_2 \leftarrow f(ima_{|D_2})$ & $T_4 \leftarrow f(ima_{|D_4})$ \\
$T_{12} \leftarrow T_1 \oplus T_2$ & $T_{34} \leftarrow T_3 \oplus T_4$ \\
\texttt{Wait thread 2} & \texttt{Wait thread 3} \\
$T_\Omega \leftarrow T_{12} \oplus T_{345}$ & $T_{345} \leftarrow T_{34} \oplus T_{5}$ \\
\end{tabular}
\end{minipage}
}
\caption{Map-reduce idiom for max-tree computation. (a) Sub-domains of
  $ima$. (b) A possible distribution of jobs by threads. (c) Map-reduce
  operations. $f$ is the map operator, $\oplus$ the merge operator.}
\label{fig:parallelize}
\end{figure}
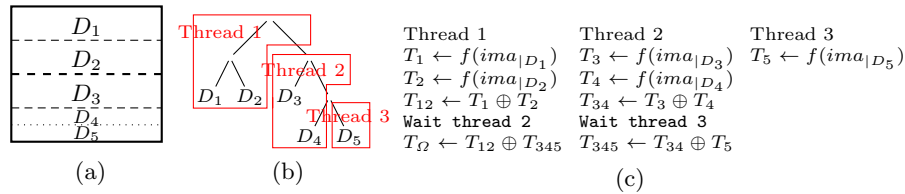

Merge-based algorithms consist in computing max-tree on sub-parts of
images and merging back trees to get the max-tree of the whole
image. Those algorithms are typically well-suited for parallelism
since they adopt a map-reduce idiom. Computation of sub max-trees (map
step), done by any sequential method and merge (reduce-step) are
executed in parallel by several threads. In order to improve cache
coherence, images should be split in contiguous memory blocks that is,
splitting along the first dimension if images are
row-major. \autoref{fig:parallelize} shows an example of parallel
processing using map-reduce idiom. Choosing the right number of splits
and jobs distribution between threads is a difficult topic that depends
on the architecture (number of threads available, power frequency of
each core). If the domain is not split enough (number of chunks no
greater than number of threads) the parallelism is not maximal, some
threads become idle once they have done their jobs, or wait for other
thread to merge. On the other hand, if the number of split gets too
large, merging and thread synchronization cause significant overheads.
Since work balancing and thread management are outside the current
topic, they are delayed to high level parallelism library such as TBB.

The procedure in charge of merging sub-trees $T_i$ and $T_j$ of two
adjacent domains $D_i$ and $D_j$ is given in algorithm
\autoref{alg:merge}. For two neighbors $p$ and $q$ in the junction of
$D_i$, $D_j$, it connects components of $p$'s branch in $T_i$ to
components of $q$'s branch in $T_j$ until a common ancestor is
found. Let $x$ and $y$, canonical elements of components to merge with
$ima(x) \ge ima(y)$ ($x$ is in the childhood to $y$) and $z$,
canonical element of the parent component of $x$. If $x$ is the root
of the sub-tree then it gets attached to $y$ and the procedure
ends. Otherwise, we traverse up the branch of $x$ to find the
component that will be attached to $y$ that is the lowest node having
a level greater than $ima(y)$. Once found, $x$ gets attached to $y$,
and we now have to connect $y$ to $x$'s old parent. Function
\texttt{findrepr(p)} is used to get the canonical element of $p$'s
component whenever the algorithm needs it.

\begin{algorithm}[htb]
\caption{Tree merge algorithm}
\label{alg:merge}
\begin{algorithmic}
\Function{findrepr}{$par$, $p$}
\If{$ima(p) \ne ima(par(p))$} \Return $p$ \EndIf
\State $par(p) \gets \Call{findrepr}{par, par(p)}$
\State \Return $par(p)$
\EndFunction
\Statex
\Procedure{connect}{p,q}
\State $x \gets \Call{findrepr}{parent, p}$
\State $y \gets \Call{findrepr}{parent, q}$
\If {$ima(x) < ima(y)$} $\Call{swap}{x,y}$ \EndIf
\While {$x \ne y$} \Comment{common ancestor found ?}
\State $parent(x) \gets \Call{findrepr}{parent, parent(x)}$;
\State $z \gets parent(x)$
\If {$x = z$} \Comment{$x$ is root}
\State $parent(x) \gets y$; $y \gets x$
\ElsIf {$ima(z) \ge ima(y)$}
\State $x \gets z$
\Else
\State $parent(x) \gets y$
\State $x \gets y$
\State $y \gets z$
\EndIf
\EndWhile
\EndProcedure
\Statex
\Procedure{mergetree}{$D_i$, $D_j$}
\ForAll {$(p,q) \in D_i \times D_j$ such that $q \in \mathcal{N}(p)$}
\State \Call{connect}{$p$,$q$}
\EndFor
\EndProcedure
\end{algorithmic}
\end{algorithm}

Once sub-trees have been computed and merged into a single tree, it
does not hold canonical property (because non-canonical elements are
not updated during merge). Also, reduction step does not merge $S$
array corresponding to sub-trees (it would imply reordering $S$ which
is more costly than just recomputing it at the end). Algorithm
\autoref{alg:canS} performs canonization and reconstructs $S$ array
from $parent$ image. It uses an auxiliary image $dejavu$ to track
nodes that have already been inserted in $S$. As opposed to other
max-tree algorithms, construction of $S$ and processing of nodes are
top-down. For any points $p$, we traverse in a recursive way its path
to the root to process its ancestors. When the
recursive call returns, $parent(p)$ is already inserted in $S$ and
holds the canonical property, thus we can safely insert back $p$ in $S$
and canonize $p$ as in algorithm \autoref{alg:canonize}.

\begin{algorithm}[htb]
\caption{Canonization and $S$ computation algorithm}
\label{alg:canS}
\begin{algorithmic}
\Procedure{CanonizeRec}{p}
\State $dejavu(p) = true$
\State $q \gets parent(p)$
\If {\textbf{not} $dejavu(q)$} \Comment{Process parent before $p$}
\State \Call{CanonizeRec}{q}
\EndIf
\If {$ima(q) = ima(parent(q)$} \Comment{Canonize}
\State $parent(p) \gets parent(q)$
\EndIf
\State \Call{InsertBack}{$S,p$}
\EndProcedure
\State
\ForAll {$p$} $dejavu(p) \gets False$ \EndFor
\ForAll {$p \in \Omega$ such that \textbf{not} $dejavu(p)$}
\State $\Call{CanonizeRec}{p}$
\EndFor
\end{algorithmic}
\end{algorithm}

\subsection{Implementation details}
Algorithms have been implemented in pure C++ using STL implementation
of some basic data structures (heaps, priority queues),
\textsc{Milena} image processing library to provide fundamental image
types and I/O functionality, and \textsc{Intel TBB} for parallelism.
Specific implementation optimizations are listed below:
\begin{itemize}
\item Sort optimization. Counting sort is used when quantization is
  lower than 18 bits. For large integer of $q$ bits, it switches to $2^16$-based
  radix sort requiring $q/16$ counting sort.
\item Pre-allocation. Queues and stacks\ldots are pre-allocated to avoid
  dynamic memory reallocation. Hierarchical queues are also
  pre-allocated by computing image histogram as a pre-processing.
\item Priority-queues. Heap is implemented with hierarchical queues
  when quantization is less than 18 bits. For large integer it switches
  to the STL standard heap implementation. A $y$-fast trie can be used
  for large integer ensuring a better complexity (see
  \autoref{sec:complexity}) but no performance gain has been
  obtained.
\item Map-reduce. In parallel version of algorithms, all instructions
  that deals about $S$ construction and $parent$ canonization have
  been removed since they are $S$ is reconstructed from scratch and
  $parent$ canonized by procedure \autoref{alg:canS}
\end{itemize}

\subsection{Complexity analysis}
\label{sec:complexity}
Let $n = H*W$ with $H$ the image height, $W$ the image width and $n$
the total number of pixels. Let $k$, the number of values in $V$.
\begin{itemize}
\item Immersion-based algorithms requires sorting pixels which has a
  complexity of $\Theta(n + k)$ ($k \ll n$) for small integers (counting sort),
  $O(n \log \log n)$ for large integers (hybrid radix sort), and $O(n \log n)$ for
  generic data type with a more complicated ordering relation
  (comparison sort). The union-find is $O(n \log n)$ and $O(n
  \alpha(n))$ when used with union-by-rank.
  \footnote{$\alpha(n)$, the inverse of Ackermann function, is very
    low growing, $\alpha(10^{80}) \simeq 4$}. The canonization step is
  linear and does not use extra memory. Memory-wise, sorting may
  require an auxiliary buffer depending on the algorithm and
  histograms for integer sorts so $\Theta(n + k)$. Union without rank
  requires a $zpar$ image for path compression $(\Theta(n))$ and the
  system stack for recursive call in \texttt{findroot} which is $O(n)$
  (\texttt{findroot} could be non-recursive, but memory space is saved
  at cost of an higher computational time). When union-by-rank is
  used, it requires two extra images ($rank$ and $repr$) of $n$ pixels
  each.
\item Flooding-based algorithms require a heap or hierarchical queues
  to retrieve the highest point in the propagation queue. Each point
  is inserted once and removed once (however they may be visited more
  than once in non-recursive versions) thus the complexity is
  $\Theta(n.p)$ where $p$ is the cost of pushing or popping from the
  heap. If the heap is encoded with a hierarchical queue as in
  \citep{salembier1998antiextensive} or \citep{nister2008linear}, it
  uses $n + 2.k$ memory space, insertion is $O(1)$, access to the
  maximum is $O(1)$ but popping is $O(k)$ (in the recursive version,
  we loop on $levroot$ to get the parent level). In practice, in images
  with small integers, gray level difference between neighboring
  pixels is far to be as large as $k$. With high dynamic image, the
  heap can be implemented as a $y$-fast trie, which has insertion and
  deletion in $O(\log \log k)$ and access to maximum element in
  $O(1)$. For any other data type $V$, a "standard" heap based on
  comparisons requires $n$ extra space, allows insertion and deletion
  in $O(\log n)$ and has a constant access to maximal element. Those
  algorithms need also an array or a stack $levroot$ to store
  canonical elements of respective size $k$ and $n$. Salembier's
  algorithm uses the system stack for a recursion of maximum depth
  $k$, hence $O(k)$.
\item For merge-based algorithm, complexity is a bit-harder to
  compute. Let $\mathcal{A}(k,n)$ the complexity of the underlying
  algorithm used to compute max-tree on sub-domains and $s=2^h$ the number
  of sub-domains. Map-reduce algorithm requires $s$ mapping operations
  and $s-1$ merges. A good map-reduce algorithm would lead split domain
  to form a full and complete tree so we assume all leaves
  to be at level $h$. Merging sub-trees of size $n/2$ has been analyzed
  in \cite{wilkinson2008concurrent} and is $O(k \log n)$ (we merge
  nodes of every $k$ levels using union-find without
  union-by-rank). Thus, the complexity of the reduction is $O(Wk \log
  n)$ and the complexity $S$ as a function of $n$ and $k$ of the
  map-reduce is:
  \begin{align*}
    S(k, i) &= \begin{cases}
      2.S(k, \frac{i}{2}) + kW \log i & \\
      \mathcal{A}(k,i) & \text{ if } i < \frac{n}{s} \\
    \end{cases}
  \end{align*}
  Assuming $s$ constant and $H = W = \sqrt{n}$ this equation solves to:
  \begin{equation*}
    S(k, n) = s.\mathcal{A}(k,\frac{n}{s}) + O(k.\sqrt{n} \log n \log s)
= O(\mathcal{A}(k,n)) + O(k\sqrt{n} \log n)
  \end{equation*}

  When there is as much splits as rows, $s$ in now dependent of $n$, it leads to
  Matas et al. \cite{matas2008parallel} algorithm whose complexity is:
  \begin{equation*}
    S(k, n) = H.W + O(k.W \log n \log H) \simeq  O(n) + O(k \sqrt{n} (\log{n})^2)
  \end{equation*}
  Contrary to what they claim, when values are small integers the
  complexity stays linear and is not dominated by merging operations.
  Finally, canonization and $S$ reconstruction has a linear time
  complexity (\texttt{CanonizeRec} is called only once for each point)
  and only uses an image of $n$ elements to track points already processed.
\end{itemize}

\begin{table}[htb]
\caption{Comparison of time complexity of many max-tree
  algorithms. $n$ is the number of pixels and $k$ the number of gray
  levels.}  \setlength{\tabcolsep}{5pt}
\begin{tabular}{l|lll}
& \multicolumn{3}{c}{Time Complexity} \\
Algorithm & Small int & Large int & Generic $V$ \\
Berger \citep{berger2007effective} & $O(n \log n)$ & $O(n \log n)$ &
$O(n \log n)$ \\
Berger + rank & $O(n.\alpha(n))$ & $O(n \log \log n)$ &
$O(n \log n)$ \\
\citet{najman2006building} & $O(n.\alpha(n))$ & $O(n
\log \log n)$ & $O(n \log n)$ \\
\citet{salembier1998antiextensive} & $O(n.k)$ & $O(n.k)
\simeq O(n^2)$ & N/A \\
\citet{nister2008linear} & $O(n.k)$ & $O(n.k) \simeq O(n^2)$ & N/A \\
\citet{wilkinson2011fast} & $O(n \log n)$ & $O(n \log n)$ & $O(n
\log n)$ \\
Salembier non-recursive & $O(n.k)$ & $O(n \log \log n)$ & $O(n \log n)$ \\
\citet{menotti20071d} (1D)  & $O(n)$ & $O(n)$ & $O(n)$ \\
Map-reduce & $O(A(k,n))$ & $O(A(k,n)) +$ & $O(A(k,n)) +$ \\
& & $O(k\sqrt{n}\log{n})$ & $O(k\sqrt{n}\log{n})$\\
\citet{matas2008parallel} & $O(n)$ & $O(n) + O(k \sqrt{n}
(\log{n})^2)$ & -
\end{tabular}
\end{table}

\begin{table}[htb]
\caption{Comparison of space requirements of many max-tree
  algorithms. $n$ is the number of pixels and $k$ the number of gray levels.}
\setlength{\tabcolsep}{5pt}
\begin{tabular}{l|lll}
& \multicolumn{3}{c}{Auxiliary space requirements} \\
Algorithm & Small int & Large int & Generic $V$ \\
\citet{berger2007effective} & $n + k + stack$ & $2n + stack$ & $n + stack$\\
Berger + rank & $3n + k + stack$ & $4.n + stack$ & $3n + stack$\\
\citet{najman2006building} & $5n + k + stack$ & $6n + stack$ & $5n +
stack$\\
\citet{salembier1998antiextensive} & $3k + n + stack$ & $2k + n + stack$ & N/A \\
\citet{nister2008linear} &  $2k + 2n$ & $2k + 2n$ & N/A \\
\citet{wilkinson2011fast} & $3n$ & $3n$ & $3n$ \\
Salembier non-recursive & $2k + 2n$ & $3n$ & $3n$ \\
\citet{menotti20071d} (1D) &  $k$ & $n$ & $n$ \\
Map-reduce & $\ldots+n$ & $\ldots+n$ & $\ldots+n$\\
\end{tabular}
\end{table}

\section{Experiments}
\begin{figure}[htb]
\subfloat[]{
  \includegraphics[width=0.5\linewidth, height=4cm]{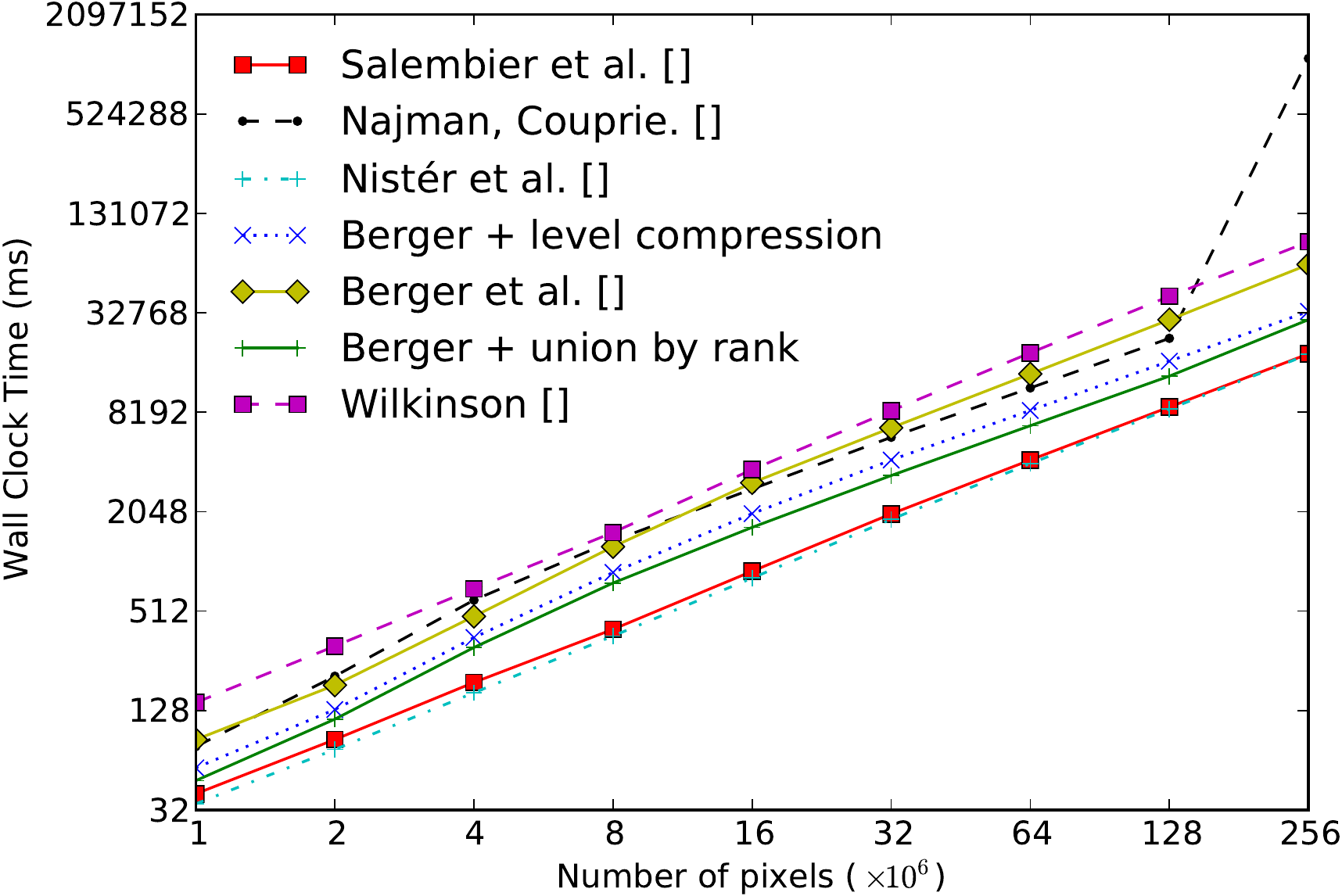}
}
\subfloat[]{
  \label{fig:cmp1b}
  \includegraphics[width=0.5\linewidth, height=4cm]{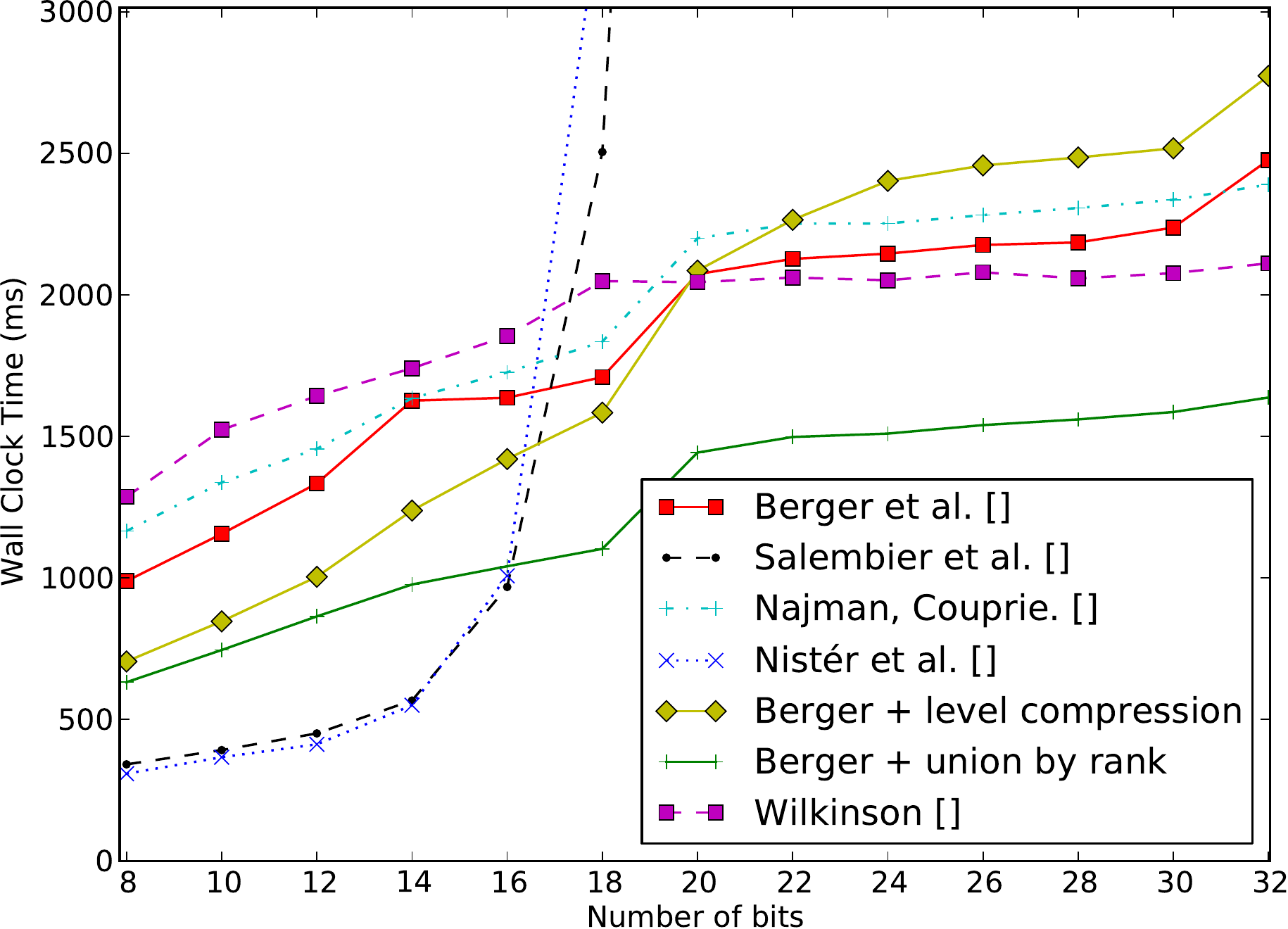}
}
\caption{ (a) Comparison of algorithms on a 8-bit image as a function
  of size; (b) Comparison of algorithms on a 6 Mega-pixels image as a
  function of quantization.}
\label{fig:cmp1}
\end{figure}

\begin{figure}[htb]
\includegraphics[width=\linewidth]{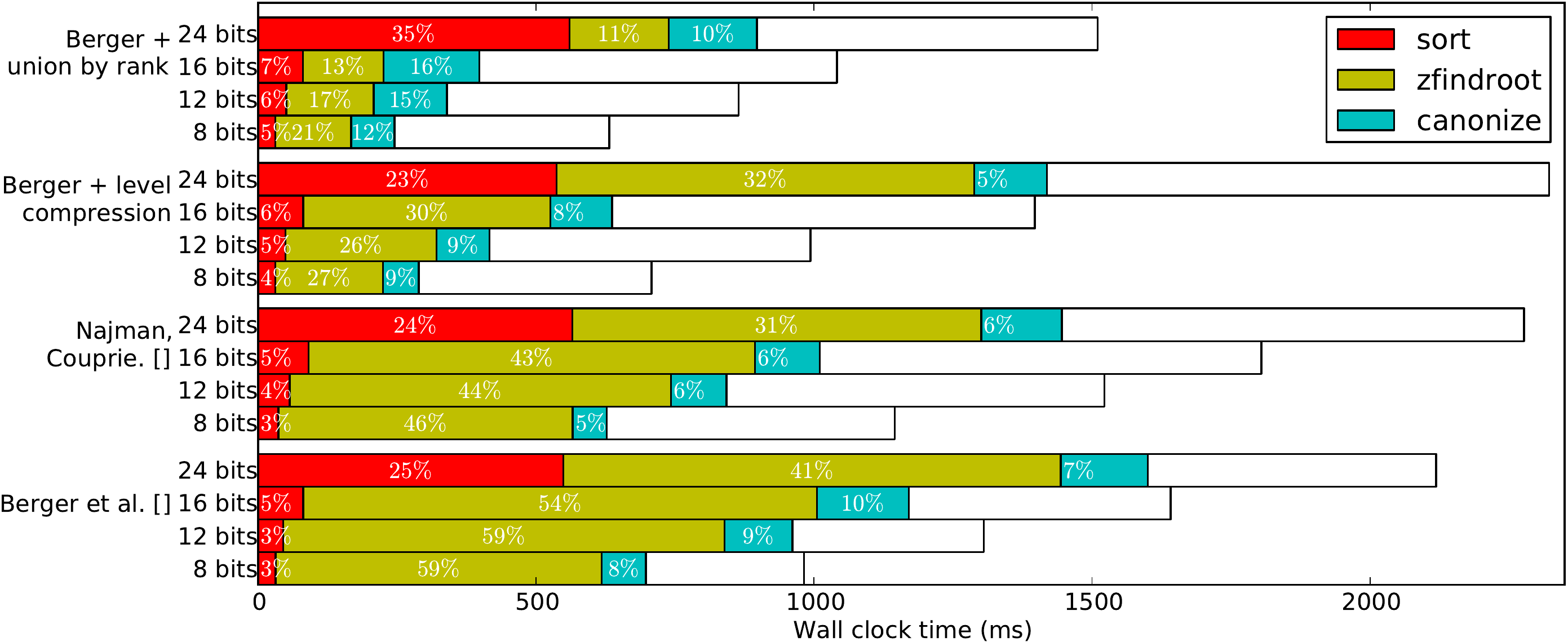}
\caption{Time spent in each part of the sequential version of
  union-find based algorithms.}
\label{fig:part}
\end{figure}

Benchmark were performed on an Intel Core i7 (4 physical cores, 8
logical cores). Program have been compiled with gcc 4.7, optimization
flags on (\texttt{--O3 --march=native}). Tests have been conducted on
a 6MB 8-bit image. Image has been resized by cropping or tiling the
original image and over-quantization has been performed by shifting
the eight bits left and generating missing lower bits at
random. \autoref{fig:cmp1} evaluates performance of sequential
algorithms w.r.t to image size and quantization. A first remark, we
notice that all algorithms are linear in practice. On natural image,
the upper bound $n \log n$ complexity of \citet{wilkinson2011fast} and
\citet{berger2007effective} algorithms is not reached. Let start with
union-find based algorithms. \citet{berger2007effective} and
\citet{najman2006building} have quite the same running time ($\pm 6\%$
on average), however performances of \citet{najman2006building}
algorithm drops significantly at 256 Mega-pixels. Indeed, at that size
each auxiliary array/image requires 1 GB memory space, thus
\citet{najman2006building} that uses much memory exceeds the 6 GB RAM
limit and needs to swap with the hard drive. Our implementation of
union-by-rank uses less memory and is on average $42 \%$ faster than
\citet{najman2006building}. Level compression is an efficient
optimization that provides $35 \%$ speed up on average on
\citet{berger2007effective}. However, this optimization is only
reliable on low quantized data, figure \autoref{fig:cmp1b} shows that
it is relevant until 18 bits. Since it operates on flat-zones, when
quantization gets higher flat-zones are less probable, thus time saved
in \Call{findroot}{} to find canonical elements does not balance
extra tests overheads (see \autoref{fig:part}. Union-find is not
affected by the quantization but sorting does, counting sort and radix
sort complexities are respectively linear and logarithmic with the
number of bits. The break in union-find curves between 18 and 20 bits
stands for the switch from counting to radix sort. Flooding-based
algorithms using hierarchical queues outperform our union-find by rank
on low quantized image by $41\%$ on average. As expected
\citet{salembier1998antiextensive} and \citet{nister2008linear} (which
is the exact non-recursive version of the first one) closely
match. However, the exponential cost of hierarchical queues w.r.t the
number of bits is evident on figure \autoref{fig:cmp1b}. By using a
standard heap instead of hierarchical queues,
\citet{wilkinson2011fast} does scale well with the number of bits and
outperform every algorithms except our implementation of
union-by-rank. In \citep{wilkinson2011fast}, the algorithm is supposed
to match \citet{salembier1998antiextensive}'s method for low quantized
images, but in our experiment it stays 4 times slower. As a
consequence:
\begin{itemize}
\item since \citet{najman2006building}'s algorithm is always
  outperformed by our implementation of union-find by rank, it will
  not be tested any further.
\item \citep{nister2008linear} and \citep{wilkinson2011fast} are
  merged in our single implementation (called \textit{Non-recursive
    Salembier} below) that will use hierarchical queues for heap when
  quantization  is below 18 and switches to a standard heap
  implementation otherwise.
\item the algorithm \textit{Berger + level compression} will enable
  level compression only when quantization is below 18 bits.
\end{itemize}

\begin{figure}[htb]
\subfloat[]{
  \includegraphics[width=0.33\linewidth, height=3cm]{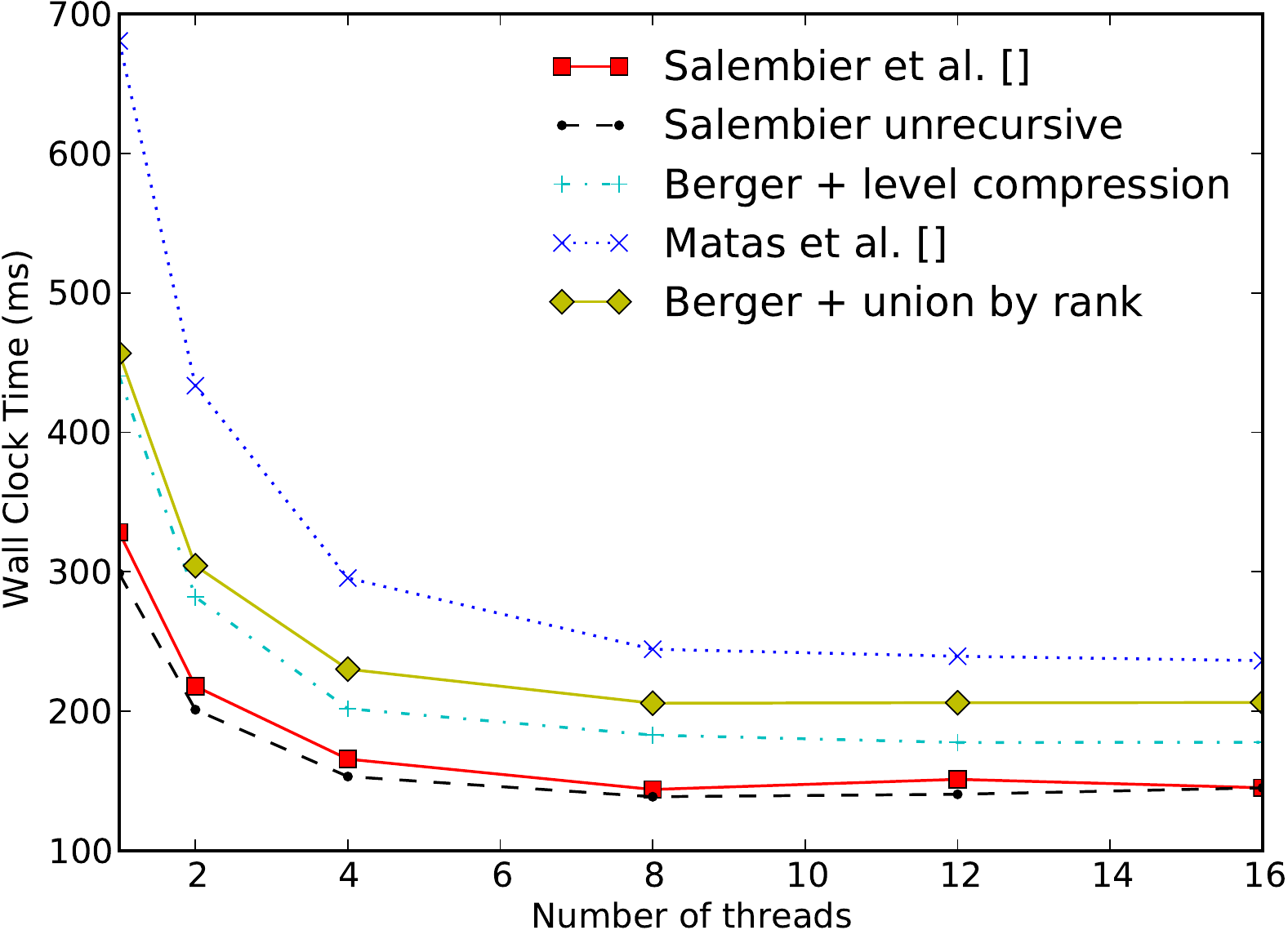}
}
\subfloat[]{
  \label{fig:cmp2b}
  \includegraphics[width=0.33\linewidth, height=3cm]{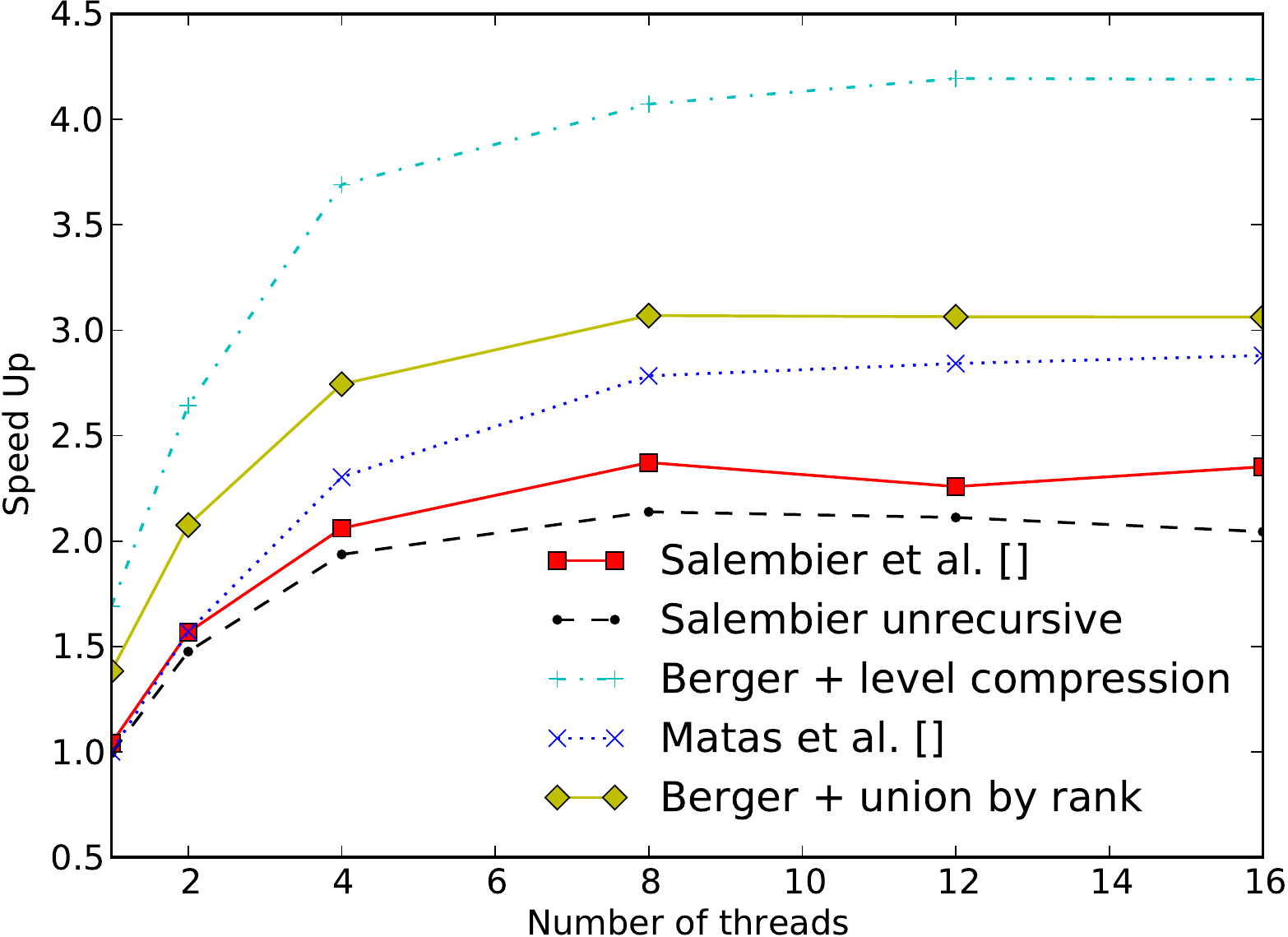}
}
\subfloat[]{
  \label{fig:cmp2c}
  \includegraphics[width=0.33\linewidth, height=3cm]{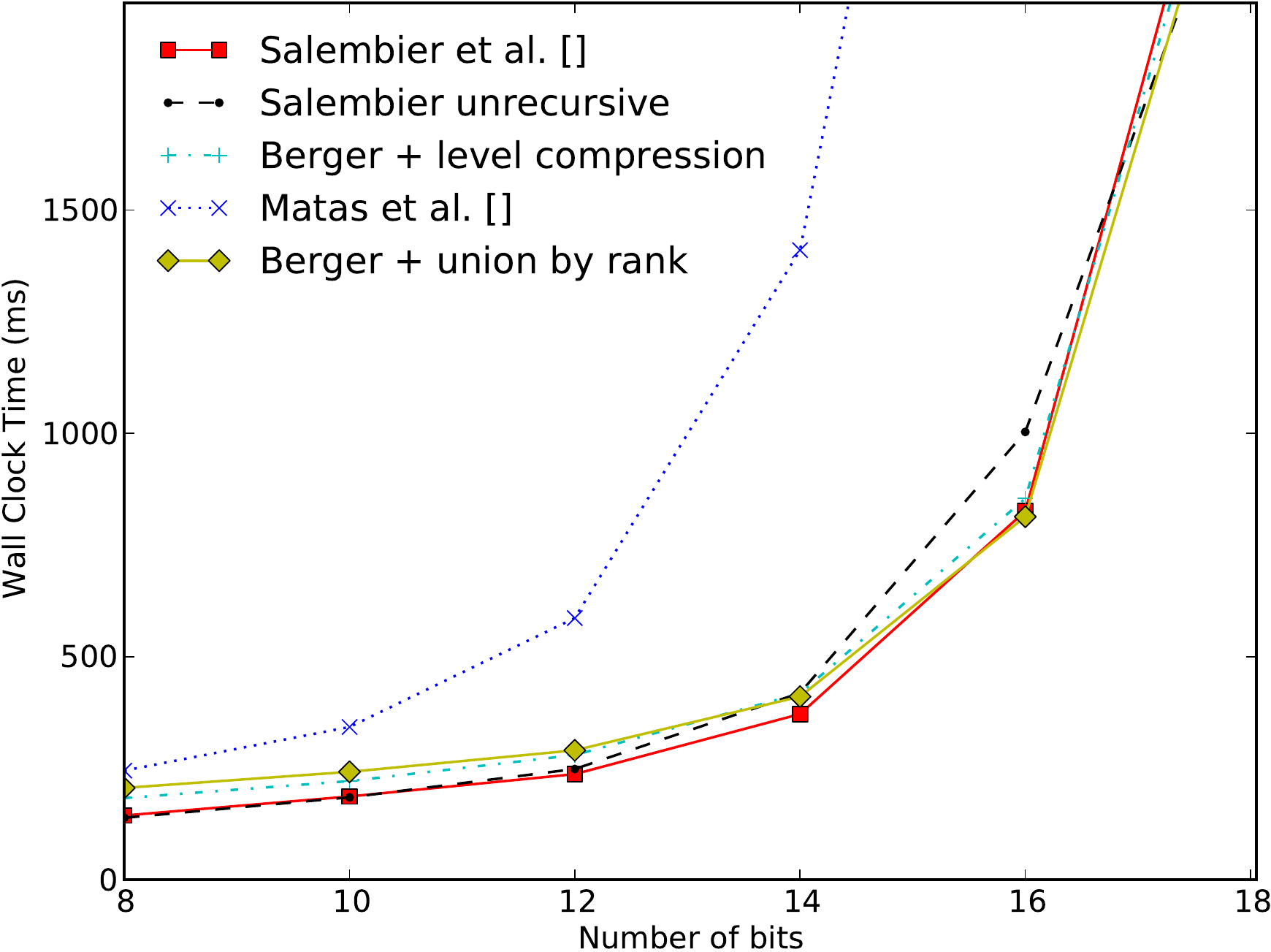}
}
\caption{(a,b) Comparison of parallel algorithms on a 6 Mega-pixels 8-bits image as a function
  of number of threads; (a) Wall clock time; (b) Speed up w.r.t the sequential
  version; (c) Comparison of parallel algorithms on a 6 Mega-pixels image as a
  function of quantization.}
\label{fig:cmp2}
\end{figure}

\begin{figure}[htb]
\includegraphics[width=\linewidth]{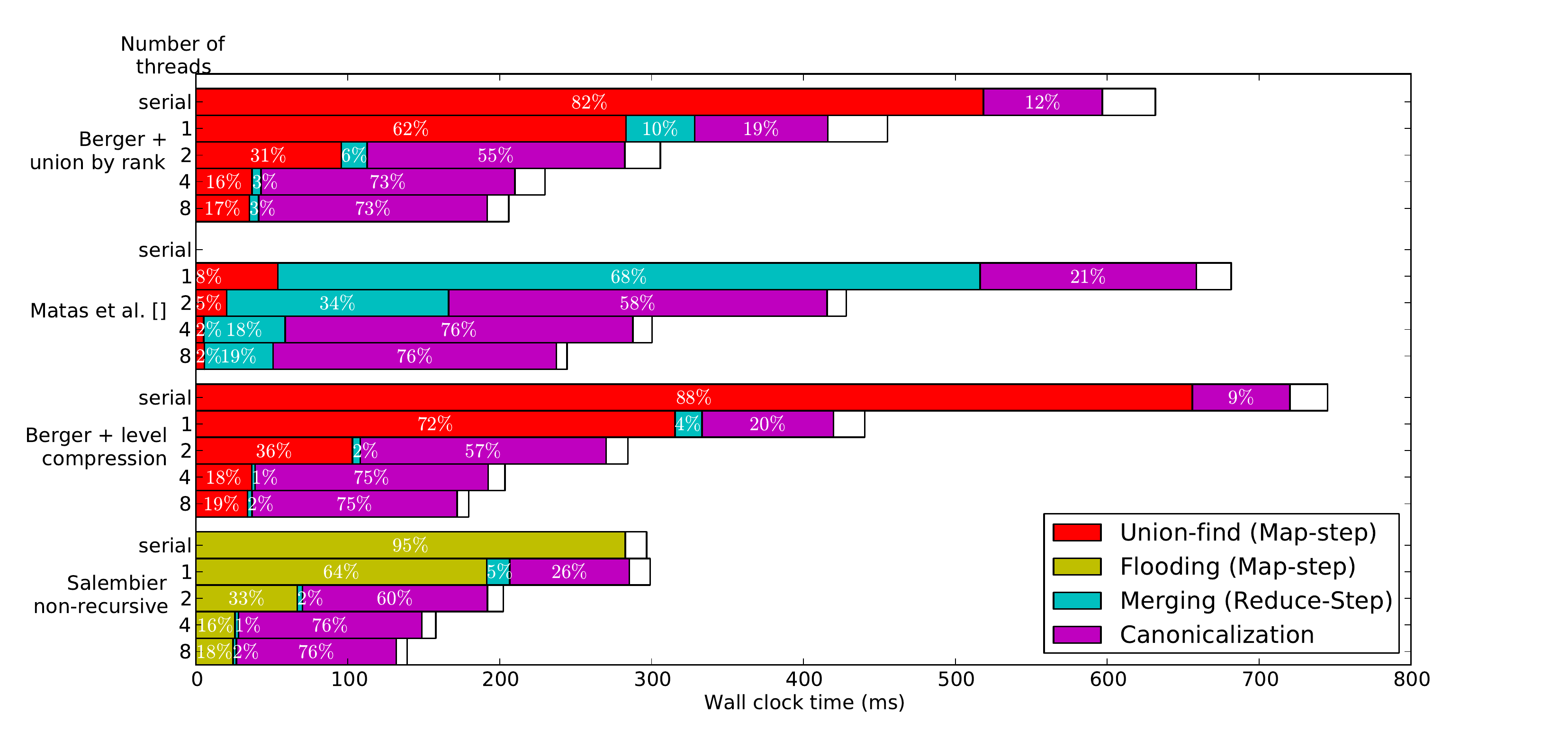}
\caption{Time spent in each part of parallel versions of algorithms.}
\label{fig:part2}
\end{figure}

\begin{figure}[htb]
\begin{tikzpicture}[]
\path[align=left, right, grow=east, level distance=80pt,
rlabel/.style={anchor=west},
level 1/.style={sibling distance=61pt, level distance=90pt},
level 2/.style={sibling distance=25pt, level distance=80pt},
parent anchor=east, child anchor=west
]
node[rlabel, align=center] {Embedded system ?\\ (memory limitation)}
	child { node[rlabel, align=center] {Low\\ quantization ?}
		child { node[rlabel] {Berger + rank}
		edge from parent node[below] {no}
		}
		child { node[rlabel] {\citet{salembier1998antiextensive} or\\ \citet{nister2008linear}\\($\pm$ parallelism)} 
		edge from parent node[above] {yes}
		}
		edge from parent node[below] {no}
	}
	child { node[rlabel, align=center] { Low\\ quantization ? } 
		child { node[rlabel] {Berger + level compression\\ ($\pm$ parallelism)}
			edge from parent node[below] {no}
		}
		child { node[rlabel] {\citet{berger2007effective}}
		edge from parent node[above] {yes}
		}
		edge from parent node[above] {yes}
	};
\end{tikzpicture}
\caption{Decision tree to choose the right max-tree algorithm}
\label{fig:decision}
\end{figure}
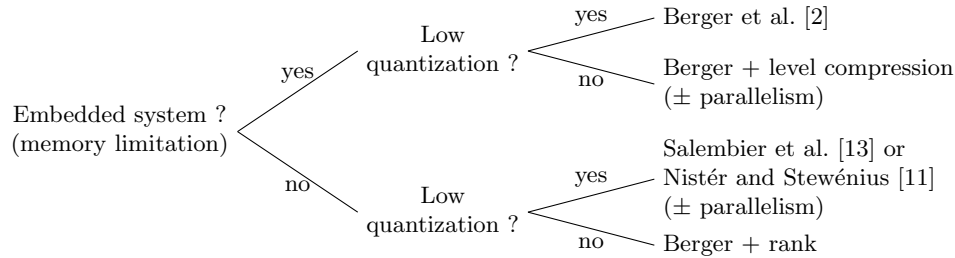

\autoref{fig:cmp2} shows results of the map-reduce idiom applied on
many algorithms and their parallelism. As a first result, we can see
best performance are generally achieve with 8 threads that is when the
number of thread matches the number of (logical) cores. However, since
there are actually only 4 physical cores, we can expect a $\times 4$
maximum speed up. Some algorithms take more benefits from map-reduce
than others. Union-find based are algorithms are particularly
well-suited for parallelism, union-find with level compression
achieves the best speed up ($\times 4.2$) at 12 threads and union-find
by rank a $\times 3.1 $ speed up with 8 threads. More surprising,
map-reduce pattern achieves significant speed up even when a single
thread is used (respectively $\times 1.7$ and $\times 1.4$ for
union-find with level compression and union-find by rank). This result
that used to surprise \citet{wilkinson2008concurrent} as well is
explained by a better cache coherence when working on sub-domains that
balances tree merges overhead. On the other hand, flooding algorithms
do not scale as well because they are limited by the canonization and
$S$ reconstruction post-process (that is going to happen also for
union-find algorithms on architecture with more cores). In
\citep{wilkinson2008concurrent} and \citep{matas2008parallel}, they
obtain a speed up almost linear with the number of threads because
only a $parent$ image is built. If we remove canonization and $S$
construction step, we also get those speed ups.(\fixme{Ajouter les
  diagrammes de rep des algo en steps}). Figure \autoref{fig:cmp2c}
shows the exponential complexity of merging trees as number of bits
increases that makes parallel algorithms unsuitable for high quantized
data. In the light of the previous analysis,  \autoref{fig:decision}
provides guidelines to choose the right max-tree algorithm w.r.t to
image types and architectures.

\section{Conclusion}
In this paper, we have tried to lead a fair comparison of max-tree
algorithms in a unique framework.  We have highlighted the fact that
there is no best algorithm that supersedes all the others in every
cases and eventually we have given a decision tree to choose the right
algorithm w.r.t to data and hardware.  We have proposed a max-tree
algorithm using union-by-rank that outperforms the existing one from
\citep{najman2006building}.  Furthermore, we have proposed a second
one that uses level compression for systems with strict memory
constraints.  As further work, we shall include image contents as a
new parameter of comparison, for instance images with large flat zones
(e.g. cartoons) or images having strongly non-uniform distribution of
gray levels. A code intensively tested used for these benachmarks is
available on the Internet at
\url{http://www.lrde.epita.fr/cgi-bin/twiki/view/Olena/maxtree}.

\bibliographystyle{splncsnat}
\bibliography{article}


\end{document}